\documentclass[10.5pt,onecolumn,pdflatex,sn-mathphys-num]{jicn}

\usepackage{graphicx}%
\usepackage{multirow}%
\usepackage{amsmath,amsthm,amssymb,amsfonts,dsfont,pifont,bm,bbm,mathrsfs,mathtools,nicefrac,extarrows,relsize}%
\usepackage[title]{appendix}%
\usepackage{textcomp}%
\usepackage{manyfoot}%
\usepackage{algorithm}%
\usepackage{algorithmicx}%
\usepackage{algpseudocode}%
\usepackage{listings}%
\usepackage{overpic}
\usepackage{wrapfig}
\usepackage{subcaption}
\usepackage[dvipsnames,table,xcdraw]{xcolor}
\usepackage{colortbl}
\usepackage{adjustbox,diagbox,tabularray}
\usepackage{multicol}

\usepackage{threeparttable}
\usepackage[utf8]{inputenc}
\usepackage{booktabs}
\usepackage{caption}
\usepackage{titlesec}
\usepackage{tablefootnote}
\usepackage{times}
\usepackage{mathptmx}
\usepackage{float}\usepackage{lastpage}
\usepackage[english]{babel}
\usepackage{setspace}
\usepackage{array}
\usepackage{makecell}
\usepackage{url}
\usepackage{xurl}
\usepackage{soul}
\def\JournalName{Journal of Intelligent Computing and Networking} 
\def\JournalAbbr{J. Intell. Comput. Netw.} 
\def\JournalURL{https://www.ffspub.com/index.php/jicn/index} 
\def\JournalEmail{jicn.office@ffspub.com} 
\def\JournalISSNprint{ 3079-9228 (print)} 
\def\JournalISSNonline{ 3107-2798 (online)} 

\def\PaperYear{2026}
\def\PaperVolume{2}
\def\PaperIssue{4}
\def\PaperStartPage{1}
\def\PaperEndPage{\pageref*{LastPage}}
\def\authorshort{Yu \textit{et al}.}
\def\PaperDOI{10.64509/jicn.24.159}
\def\PaperCopyrightOwner{2026 The authors. This article is an open access article distributed under the terms and conditions of the Creative Commons Attribution (CC BY) license (https://creativecommons.org/licenses/by/4.0/).}

\usepackage{geometry}
\usepackage{fancyhdr}
\usepackage[hang,flushmargin]{footmisc}

\titleformat{\section}
  {\normalfont\fontsize{16}{16}\bfseries}
  {\thesection}{1em}{}
\titlespacing{\section}{0pt}{1em}{1em}
\titleformat{\subsection}
  {\normalfont\fontsize{12}{12}\bfseries}
  {\thesubsection}{1em}{}
\titlespacing{\subsection}{0pt}{0.8em}{0.8em}
\titleformat{\subsubsection}
  {\normalfont\fontsize{10}{10}\bfseries}
  {\thesubsubsection}{1em}{}
\titlespacing{\subsubsection}{0pt}{0.5em}{0.5em}

\def\journalinfohead{\authorshort{} / \JournalAbbr{} \quad \PaperYear{} \PaperVolume(\PaperIssue):\PaperStartPage--\PaperEndPage}
\makeatletter
\def\blfootnote#1{\gdef\@thefnmark{}\@footnotetext{#1}}
\makeatother
\renewcommand{\footnoterule}{\kern -3pt \hrule width 0.3\columnwidth \kern 2.6pt}

\hypersetup{
  colorlinks=true,
  urlcolor=blue,
  citecolor=blue,
  linkcolor=blue
}

\theoremstyle{thmstyleone}

\theoremstyle{thmstyletwo}

\theoremstyle{thmstylethree}

\ifdefined\ReviewVersion
  \newcommand{\rev}[1]{{\color{red}#1}}
  \newenvironment{revision}{\color{red}}{}
\else
  \newcommand{\rev}[1]{#1}
  
\fi
\makeatletter

\expandafter\def\csname changedbib@ni2026genvidbench\endcsname{1}
\expandafter\def\csname changedbib@chen2025genworld\endcsname{1}
\expandafter\def\csname changedbib@jiang2025ivyfake\endcsname{1}
\expandafter\def\csname changedbib@zheng2025d3\endcsname{1}
\expandafter\def\csname changedbib@BERT\endcsname{1}
\expandafter\def\csname changedbib@frank2020leveraging\endcsname{1}
\expandafter\def\csname changedbib@zeroscope1\endcsname{1}
\expandafter\def\csname changedbib@Morphstudio\endcsname{1}
\expandafter\def\csname changedbib@OpenSora\endcsname{1}
\expandafter\def\csname changedbib@videofusion\endcsname{1}
\expandafter\def\csname changedbib@Gen2\endcsname{1}
\makeatother

\newcommand{\cl}[1]{\textcolor[RGB]{70,113,184}{#1}}

\renewcommand{\arraystretch}{1.3}
\begin{document}

\noindent
\setcounter{page}{1}
\begin{minipage}[c][2.5cm][c]{0.75\textwidth}
	\fontsize{9}{11}\selectfont
	\raggedright
	\textbf{\JournalName} \\
	\JournalURL \\
	ISSN \JournalISSNprint \\
	ISSN \JournalISSNonline\\
	E-mail: \JournalEmail
\end{minipage}%
\hfill
\begin{minipage}[c][2.5cm][c]{0.2\textwidth}
  \flushright
  \includegraphics[width=3.5cm]{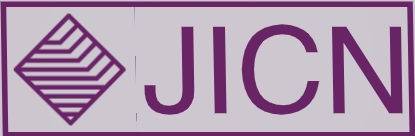}
\end{minipage}

\fontsize{10}{12}\selectfont
\noindent\textit{Article}
\begin{center}
  \fontsize{20}{24}\bfseries
  Semantic-aware Contrasting with Explicitly Disentangled Representations for AI-Generated Video Detection
\end{center}

\begin{center}
  \fontsize{11}{14}\selectfont
  Jiaquan Yu\textsuperscript{1},
  Jingyi Xie\textsuperscript{1},
  Zeyu Zhang\textsuperscript{1},
  Yunhang Qian\textsuperscript{2},
  Xiaobin Hu\textsuperscript{2},
  Jiawei Liu\textsuperscript{1,\textdagger}
  \\
  \fontsize{9}{11}\selectfont\itshape
  \textsuperscript{1}MoE Key Lab of BIPC, University of Science and Technology of China, Hefei, China, 230027\\
  \textsuperscript{2}School of Computing, National University of Singapore, Singapore, 119077\\
  E-mail: xk20031012@mail.ustc.edu.cn; hsfzxjy@mail.ustc.edu.cn; zeyuzhang10@mail.ustc.edu.cn\\
  3180655398@qq.com; ben0xiaobin0hu1@nus.edu.sg; jwliu6@ustc.edu.cn
  \\
 \vspace{1em}
 \fontsize{9}{11}\selectfont
 Received: September 18, 2026 / Revised: November 3, 2026 / Accepted: November 6, 2026
\end{center}
\vspace{1em}

\noindent\textbf{Abstract:} The rapid proliferation of AI-generated videos (AIGVs) has raised critical concerns about content authenticity and trustworthy media consumption, underscoring the need for robust AIGV detection. Despite recent progress, existing benchmarks suffer from a fundamental limitation: real and fake videos are often collected or generated from different semantic distributions. As a result, detectors may rely on spurious correlations involving objects, scenes, or actions rather than intrinsic artifacts introduced by generative models. This limitation calls for a benchmark that explicitly aligns real and generated videos at the semantic level, enabling detection methods to be evaluated and optimized under controlled content conditions. To this end, we introduce SAVGE, a large-scale semantically aligned AIGV detection dataset comprising 2,369,822 videos, including 424,115 real videos and 1,945,707 generated videos. SAVGE organizes samples into real--fake semantic groups by using captions paired with real videos as prompts for text-to-video generation, thereby providing both group-level correspondence and textual semantic anchors. Based on SAVGE, we further propose Explicit Disentangling and Semantic-aware Contrasting (EDSC), a model-agnostic training framework for learning artifact-focused representations with reduced semantic interference. EDSC first extracts and aligns visual semantic features with textual representations, and then explicitly suppresses semantic information in the artifact branch while contrasting real and fake samples within each semantic group. By integrating caption-guided semantic supervision with intra-group contrastive learning in a two-phase optimization strategy, EDSC encourages detectors to capture generation-related cues under matched semantic conditions. Extensive experiments on SAVGE and GenVideo demonstrate that EDSC \rev{improves most evaluated backbone--metric combinations}, enhances robustness under challenging evaluation settings, and strengthens cross-dataset generalization. These results highlight the importance of semantic alignment in AIGV detection and validate explicit semantic--artifact disentanglement as an effective strategy for robust detector learning.
\vspace{1em}

\href{https://doi.org/\PaperDOI}{https://doi.org/\PaperDOI}
\vspace{8mm}

\blfootnote{\textsuperscript{\textdagger} Corresponding author: Jiawei Liu (\href{mailto:jwliu6@ustc.edu.cn}{jwliu6@ustc.edu.cn}).}
\blfootnote{\textcopyright\ \PaperCopyrightOwner}

\begin{multicols}{2}

\section{Introduction}

\noindent Recent advances in generative modeling have made it increasingly practical to synthesize high-fidelity videos with realistic appearance, motion, and temporal coherence~\cite{mmgt,wang2025uniadapter,ho2022imagen,ni2023conditional,jiang2023videobooth,wang2023modelscope,expressive,denoisingreuse}. These AI-generated videos (AIGVs) are expanding the creative possibilities of entertainment, media production, education, and virtual environments. At the same time, their growing realism and accessibility raise serious concerns about misinformation, impersonation, and the erosion of trust in digital content. As synthetic videos become increasingly difficult for humans to distinguish from camera-captured ones, robust AIGV detection is becoming increasingly important for content authentication.

A central challenge in AIGV detection is distinguishing generation artifacts from semantic content. Ideally, a detector should determine whether a video is real or generated without relying on the objects, scenes, or actions it depicts. This requirement makes dataset construction particularly important: when real and fake samples differ substantially in semantics, models may learn spurious content biases rather than intrinsic traces left by generative models~\cite{ucf}. Therefore, a strong benchmark should not only contain diverse synthetic videos but also control the semantic gap between real and fake samples.

\begin{figure}[H]
	\centering
	\includegraphics[width=0.9\linewidth]{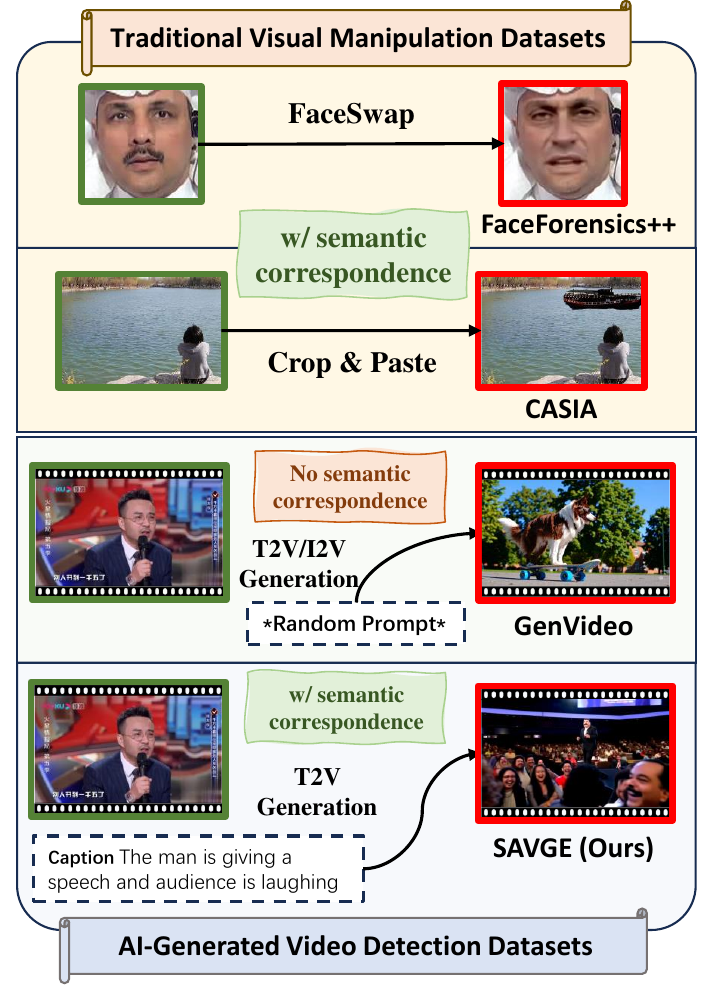}
	\caption{Comparison of representative \rev{visual manipulation} and AIGV detection datasets. FaceForensics++~\cite{roessler2019faceforensicspp} and CASIA~\cite{casia} preserve semantic correspondence because \rev{fake content is produced by manipulating real content. FaceForensics++ contains videos, whereas CASIA contains images and is included only as an image-tampering example.} By contrast, existing AIGV benchmarks such as GenVideo~\cite{DeMamba} generate fake videos from independent prompts, leading to weak semantic alignment. SAVGE addresses this gap by synthesizing fake videos from captions paired with real videos.}
	\label{fig:title}
\end{figure}

Chen \textit{et al.}~\cite{DeMamba} took an important step toward this goal by introducing GenVideo~\cite{DeMamba}, the first large-scale benchmark for AIGV detection. GenVideo contains diverse, high-quality fake videos synthesized by multiple generation techniques and defines practical evaluation tasks such as cross-generator video classification and degraded video classification. These designs make GenVideo a valuable benchmark for studying detector robustness under realistic conditions.
Nevertheless, GenVideo~\cite{DeMamba} still lacks explicit semantic correspondence between real and fake samples. As illustrated in Figure~\ref{fig:title}, \rev{visual manipulation datasets} such as FaceForensics++~\cite{roessler2019faceforensicspp} and CASIA~\cite{casia} generate fake \rev{content} by directly manipulating \rev{real content}, naturally preserving strong semantic alignment across the two domains. In contrast, \rev{GenVideo does not systematically provide the caption-based real--fake groups used in SAVGE}. This semantic mismatch obscures the intended detection signal: models may exploit content-distribution differences rather than learn generation-related artifacts. It also limits the use of learning strategies, such as explicit semantic disentanglement, that require semantically aligned supervision.
\vspace{-1em}

To bridge this gap, we introduce SAVGE, \rev{a} large-scale AIGV detection dataset built around semantically aligned real--fake video groups. SAVGE treats semantic correspondence as a core dataset property: \rev{each training group contains} one real video and multiple fake videos that share the same semantic description. Specifically, real videos are collected from existing video-captioning datasets, and their captions are used as prompts for state-of-the-art text-to-video generation models to synthesize corresponding fake videos. This design converts the original video captions from auxiliary annotations into explicit semantic anchors that connect real and generated samples at the group level. To make this correspondence reliable, we further apply a retrieval-based ranking and filtering strategy that compares generated candidates with their paired real videos, removes severely misaligned samples, and retains semantically consistent fake videos. After this filtering process, SAVGE contains 424,115 real videos and 1,945,707 fake videos, making it both large in scale and structurally aligned for controlled AIGV detection research. The resulting organization enables comparisons between real and fake videos under matched semantic conditions, reduces the risk that detectors exploit content-distribution shortcuts, and supports fine-grained analyses of generation artifacts across different semantic contexts. Moreover, the paired captions and group identifiers make SAVGE suitable for learning strategies that require semantic supervision, such as semantic--artifact disentanglement and group-level contrastive learning.

Building on SAVGE, we further propose a model-agnostic Explicit Disentangling and Semantic-aware Contrasting (EDSC) training framework for robust AIGV detection. EDSC decomposes video representations into semantic and artifact components and encourages the artifact component to be discriminative while remaining insensitive to semantic variation. It operates through a two-phase training strategy. In the Semantic Extraction and Alignment (SEA) phase, EDSC uses captions as semantic anchors and aligns the visual semantic component with textual representations by maximizing cross-modal mutual information. In the Semantic Suppression and Intra-Group Contrasting (SSGC) phase, EDSC suppresses semantic information in the artifact component by minimizing intra-modal mutual information, while a group-level contrastive objective separates real and fake videos within each semantic group. Through this alternating process, EDSC learns semantically invariant representations that better capture generation artifacts. Extensive experiments on SAVGE and GenVideo validate the effectiveness and generalizability of EDSC, and further demonstrate the value of semantic alignment for robust AIGV detection.

Our main contributions are summarized as follows:
\begin{itemize}
\item We introduce SAVGE, \rev{a} large-scale AIGV detection dataset that organizes real and fake videos into semantically aligned groups, with textual captions serving as semantic anchors.
\item We design a scalable caption-guided generation and filtering pipeline that uses real-video captions as prompts, ranks candidate generated videos, and filters them to construct reliable semantically aligned real--fake groups.
\item We propose EDSC, a model-agnostic framework that learns semantically invariant yet discriminative artifact representations through explicit disentanglement learning and semantic-aware contrasting learning.
\item We conduct extensive experiments on the SAVGE and GenVideo datasets, demonstrating the effectiveness and generalizability of EDSC and highlighting the importance of semantic alignment for robust AIGV detection.
\end{itemize}

\section{Related Work}

\subsection{Video Generation Methods}

\noindent Early video generation studies were largely driven by Generative Adversarial Networks (GANs), which extend image-level adversarial synthesis to the spatiotemporal domain by jointly modeling appearance and motion. Representative methods include TGAN~\cite{saito2017tgan}, which introduces a temporal generator to capture latent dynamics, MoCoGAN~\cite{tulyakov2018mocogan}, which disentangles motion and content representations to improve temporal coherence, and StoryGAN~\cite{li2019storygan}, which incorporates structured story-level conditioning to encourage semantic consistency across frames. These GAN-based models established important foundations for video synthesis and can produce visually sharp clips, but they are often limited by training instability, restricted temporal range, and weak semantic controllability.

Recent progress has therefore shifted toward diffusion-based and transformer-enhanced generation pipelines that support prompt-conditioned, scalable video synthesis~\cite{mmgt,wang2025uniadapter,ho2022imagen,ni2023conditional,jiang2023videobooth,wang2023modelscope,videofusion,ma2024latte,OpenSora}. Compared with earlier adversarial approaches, diffusion models provide a more stable generative framework and can better exploit textual or visual conditions to control video semantics. For example, Imagen Video~\cite{ho2022imagen} demonstrates high-definition text-to-video generation with cascaded diffusion models, VideoBooth~\cite{jiang2023videobooth} incorporates image prompts for efficient personalized generation, ModelScopeT2V~\cite{wang2023modelscope} introduces spatiotemporal modules to improve frame consistency and motion smoothness, and VideoFusion~\cite{videofusion} decomposes the diffusion process to better capture video structure. More recent systems, such as Latte~\cite{ma2024latte} and OpenSora~\cite{OpenSora}, further integrate attention-based architectures to strengthen long-range temporal dependency modeling and improve scalability.

Overall, progress in video generation has lowered the barrier to producing high-quality synthetic videos and increased the difficulty of distinguishing them from real-world content. Accordingly, this paper focuses on AI-generated video detection and constructs SAVGE using multiple text-based video generation methods to synthesize fake videos whose semantics are aligned with real video captions.

\subsection{AI-Generated Visual Content Detection}

\noindent Detection of AI-generated visual content has evolved from early GAN-artifact detectors based on low-level statistics and CNN features~\cite{wang2020cnn,durall2020watching} to methods using spatiotemporal cues and hybrid pixel-frequency representations~\cite{li2018eyeblinking,bertasius2021timesformer,frank2020leveraging}.
Although effective, these methods often depend on generator-specific artifacts and may degrade under changes in synthesis models, compression pipelines, or data distributions.
This issue is further amplified by diffusion-based generation, whose high fidelity, semantic controllability, and artifact patterns differ from earlier GAN-generated content~\cite{wang2023genimage,ojha2023universal}.
Related image-forensics research has likewise emphasized learning generic forgery cues across source qualities through multi-prior fusion~\cite{zhang2026forgery}.

Reliable benchmarks are therefore essential for assessing whether detectors learn generalizable generation cues rather than dataset-specific biases.
Representative deepfake benchmarks, including FaceForensics++~\cite{roessler2019faceforensicspp}, Celeb-DF~\cite{li2020celebdf}, and DFDC~\cite{dolhansky2020deepfake}, provide manipulated facial videos across diverse compression levels, synthesis qualities, and real-world conditions.
Because they focus mainly on face manipulation, fake samples are typically produced by modifying real source videos and thus retain strong correspondence in identity, scene context, and motion.
While useful for studying manipulation artifacts, this setting does not fully reflect AI-generated video (AIGV) detection, where an entire video can be synthesized from a text prompt and real--fake semantic distributions may be mismatched.

To support AIGV detection, Chen \textit{et al.} introduced GenVideo~\cite{DeMamba}, the first large-scale benchmark in this area, which provides diverse generated videos and practical evaluation protocols.
However, its generated videos are not explicitly paired with semantically corresponding real videos, creating a semantic distribution gap that may encourage detectors to exploit content biases rather than intrinsic generation artifacts.
To address this limitation, we introduce SAVGE, a large-scale AIGV detection dataset organized around semantically aligned real--fake video groups with textual annotations.
Using captions as semantic anchors, SAVGE enables controlled evaluation and supports learning strategies that disentangle semantic information from artifact-sensitive representations.

\rev{Recent benchmarks include GenVidBench~\cite{ni2026genvidbench}, which studies cross-source and cross-generator detection and includes real-video-derived test pairs; GenWorld~\cite{chen2025genworld}, which examines generated real-world simulation videos; and Ivy-Fake~\cite{jiang2025ivyfake}, which supports explainable image and video detection. D3~\cite{zheng2025d3} detects generated videos using second-order temporal features without detector training. SAVGE combines caption-based semantic groups with retrieval filtering and uses this structure directly in EDSC training.}

\subsection{Feature Disentanglement Learning for AI-Generated Visual Content Detection}

\noindent Feature disentanglement learning aims to separate visual representations into interpretable factors, making it appealing for forgery detection, where models should focus on manipulation artifacts rather than semantic content. Otherwise, detectors may rely on objects, scenes, or motions as shortcuts and generalize poorly to unseen generators or real-world data.
Prior work~\cite{selfsupdis,disentanglerep,fdml,ucf} has shown the value of this idea for AI-generated visual content detection. UCF~\cite{ucf} improves cross-domain face forgery detection by disentangling shared semantics from forgery cues, while FDML~\cite{fdml} combines disentanglement with multi-view learning to mine intrinsic forgery traces. However, extending such strategies to AIGV detection remains difficult since existing datasets often lack semantic correspondence between real and fake videos, making generation artifacts hard to separate from content differences.

SAVGE addresses this issue by organizing real and fake videos into semantically aligned groups and providing captions as semantic anchors. This structure enables our EDSC framework to disentangle semantic and artifact representations and apply semantic-aware contrasting within each group, encouraging semantically invariant yet artifact-sensitive features.

\section{SAVGE: Semantically Aligned AI-Generated Video Dataset}

\noindent SAVGE is a large-scale dataset for AI-generated video detection, comprising 2,369,822 videos in total, including 424,115 real videos sourced from multiple real-world datasets and 1,945,707 AI-generated fake videos. 
Unlike GenVideo~\cite{DeMamba}, which lacks explicit semantic alignment between real and fake videos, SAVGE organizes samples into semantic groups. \rev{Each group contains one real video and its caption, together with four to six fake videos for training or one fake video for testing. The caption serves as the semantic anchor.} \rev{This structure provides direct supervision for learning with captions and for contrasting real and generated videos within the same semantic group.} 
SAVGE is intended to support research on robust AI-generated video detection and related tasks.

\rev{SAVGE and GenVideo include several common generation methods, but differ in their source videos, sample organization, and filtering. We therefore evaluate the same SAVGE-trained detectors on both test sets, as described in Section~\ref{sec:exp}.} At the same time, we recognize the rapid evolution of generative techniques. Therefore, we plan to update SAVGE on an annual basis by incorporating newly emerged generation methods, keeping the benchmark comprehensive, current, and representative of real-world challenges.

\subsection{Video Collection and Generation}

\noindent To obtain high-quality textual annotations for each semantic group, SAVGE uses several video-captioning datasets as sources of real videos, including Youku-mPLUG~\cite{Youku-mPLUG}, ActivityNet~\cite{krishna2017dense}, and 20BN-s-s V2~\cite{goyal2017something}. These datasets provide a rich variety of real-world video content across multiple domains and scenarios.
We carefully select a subset from each dataset to ensure diverse semantics and video styles, forming the base set of positive samples in SAVGE.

Each real video serves as the basis of a unique semantic group and is treated as the sole positive sample in that group. To construct semantically aligned negative samples, we use the caption of each real video as the prompt for text-to-video generation. Specifically, SAVGE employs a diverse set of state-of-the-art text-to-video models to synthesize fake videos, including ZeroScope~\cite{zeroscope1}, VideoCrafter~\cite{chen2023videocrafter1}, Wan2.1~\cite{wan2025}, SD~\cite{zhang2023pia}, Latte~\cite{ma2024latte}, OpenSora~\cite{OpenSora}, ModelScope~\cite{wang2023modelscope}, MorphStudio~\cite{Morphstudio}, Show\_1~\cite{zhang2023show}, Gen2~\rev{\cite{Gen2}}, Crafter~\cite{chen2023videocrafter1}, and Lavie~\cite{Lavie}. As shown in Table~\ref{tab:dataset}, \rev{the development and test splits each use six generation sources.} Formally, we denote the $i$-th real video as $V^{i,0}$ and its generated fake videos as $\{V^{i,j}\}_{j=1}^6$, where $j$ indexes the six generation sources used for the corresponding split.

\rev{ZeroScope uses 576w generation at $576\times320$ followed by XL upscaling to $1024\times576$; Table~\ref{tab:dataset} lists the final output resolution.}

\end{multicols}
\begin{center}
\begin{threeparttable}[]
	\vspace{-2em}	
	\centering
	\caption{Statistics of real and generated videos in our proposed SAVGE dataset.}\label{tab:dataset}
	\setlength{\tabcolsep}{6pt}
	\renewcommand{\arraystretch}{1.1}
	\begin{tabular}{l|c|ccc|c|c|c}
		\toprule
		\textbf{Video Source} & \textbf{Type} & \textbf{Resolution} & \textbf{FPS} & \textbf{Length} & \textbf{Train Set} & \textbf{Test Set} & \textbf{Total Count} \\
		\midrule
		ActivityNet~\cite{krishna2017dense} &\multirow{3}{*}{Real} &-&-&\textasciitilde 120s&  13,289 & 8,090 & \multirow{3}{*}{424,115} \\
		Youku-mPLUG~\cite{Youku-mPLUG} & &-& - &10-120s& 176,510 & 5,379 &  \\
		20BN-s-s V2~\cite{goyal2017something} &  &-&-&\textasciitilde 15s& 214,316 & 6,531 &\\
		\midrule
		ZeroScope~\cite{zeroscope1} & \multirow{6}{*}{Fake}& 1024$\times$576&8& 3s& 190,605 & - & \multirow{6}{*}{1,925,707} \\
		VideoCrafter~\cite{chen2023videocrafter1} &  &1024$\times$576&8& 2s& 291,082 & - &  \\
		Wan2.1~\cite{wan2025} & & $1280\times 720$ & 16& 5s & 337,192 & - & \\
		SD \cite{zhang2023pia}& &512-1024&8&2-6s& 298,661 & - &  \\
		Latte~\cite{ma2024latte} & &512$\times$512&8&2s& 405,655 & - &  \\
		OpenSora~\cite{OpenSora}& &512$\times$512&8&2s& 402,512 & - &  \\
		\midrule
		ModelScope~\cite{wang2023modelscope} & \multirow{6}{*}{Fake} &256$\times$256&8&4s& - & 2,451 & \multirow{6}{*}{20,000} \\
		MorphStudio~\cite{Morphstudio} & &1280$\times$720&8&2s& - & 2,983 & \\
		Show\_1~\cite{zhang2023show} & &576$\times$320&8&4s& - & 3,913 &  \\
		Gen2~\rev{\cite{Gen2}} & &896$\times$512&24&4s& - & 3,246 &  \\
		Crafter~\cite{chen2023videocrafter1} & &256$\times$256&8&4s& - & 3,089 &  \\
		Lavie~\cite{Lavie} & &1280$\times$2048&8&2s& - & 4,318 &  \\
		\midrule
		\textbf{Total Count} & - &-&-&- & 2,329,822 & 40,000 & 2,369,822 \\
		\bottomrule
	\end{tabular}
	
\end{threeparttable}
\end{center}
\begin{multicols}{2}

\subsection{Ranking and Filtering for Fake Videos}

\noindent Because current text-to-video models do not always preserve prompt semantics faithfully, some generated videos may deviate from their corresponding real videos in subject, scene, or action. To reduce such mismatches, we introduce a video--video retrieval ranking mechanism to identify and discard low-quality generated samples.

Specifically, for the $i$-th semantic group with generated videos $\{V^{i,j}\}_{j=1}^6$, we adopt VideoMAE V2~\cite{wang2023videomae} to compute a similarity score $s_j \in [-1,1]$ between the real video $V^{i,0}$ and each fake sample $V^{i,j}$. \rev{We first remove groups whose mean similarity over all six candidates is below 0.50, then filter candidates within the remaining groups.} For the training set, we retain fake videos whose retrieval score is greater than 0.6 or whose rank is among the top four within the group, balancing semantic consistency with sample diversity. \rev{The top-four rule maintains the group size for training, while the score threshold admits additional well-matched candidates.} For the test set, we apply a stricter criterion and keep only the top-ranked fake video in each group. After filtering, SAVGE contains 2,369,822 videos in total, including 424,115 real and 1,945,707 fake ones.

\begin{figure}[H]
	\includegraphics[width=\linewidth]{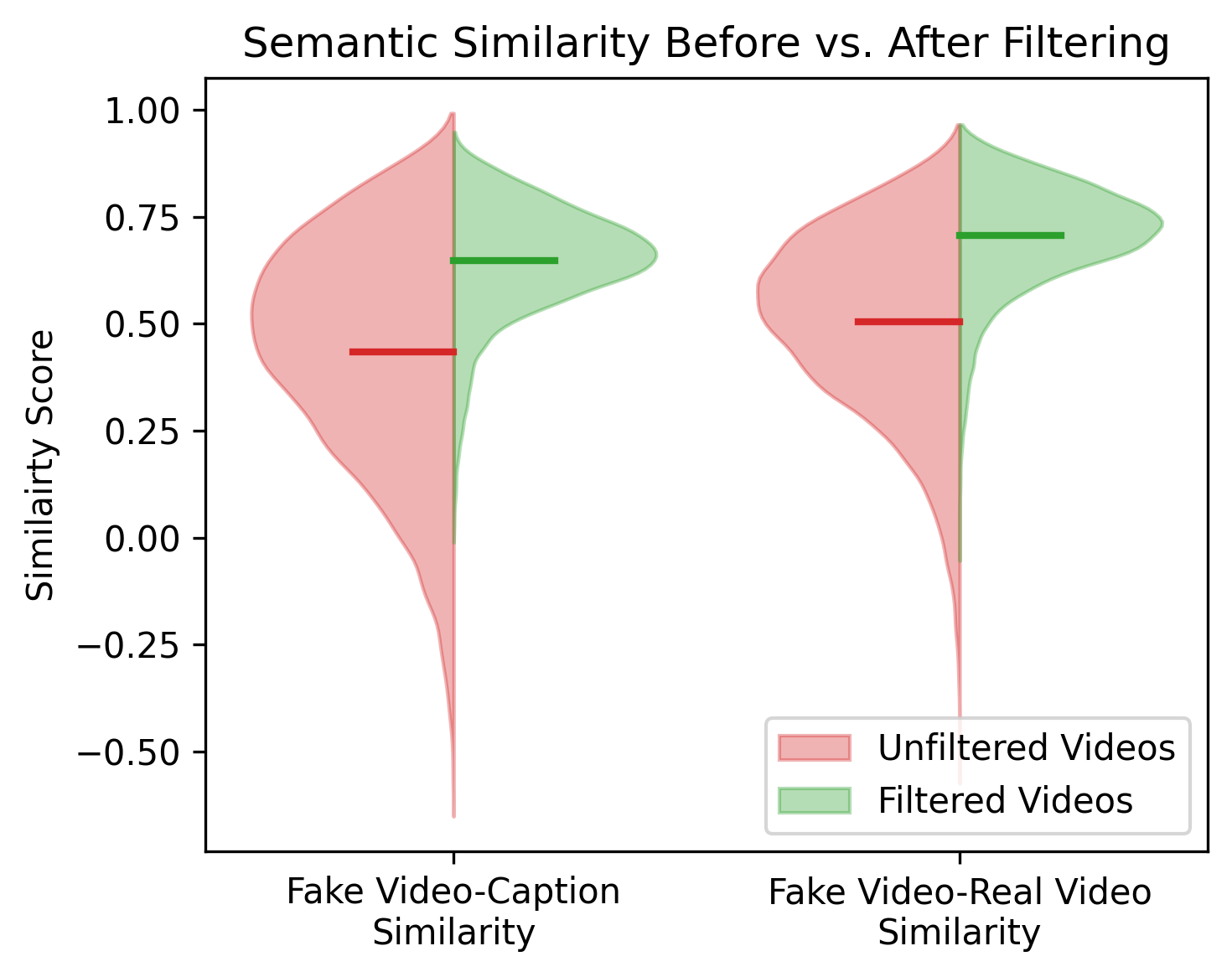}
	\caption{``Fake Video--Caption'' and ``Fake Video--Real Video'' similarity-score distributions before and after the filtering process. The distributions show that retained fake videos are more semantically aligned with their corresponding real videos or captions than \rev{videos in the unfiltered set}, indicating that the filtering process reduces real--fake semantic mismatch.\label{fig:sim}}
\end{figure}

\begin{figure*}[]
	\centering
	\includegraphics[width=0.9\textwidth]{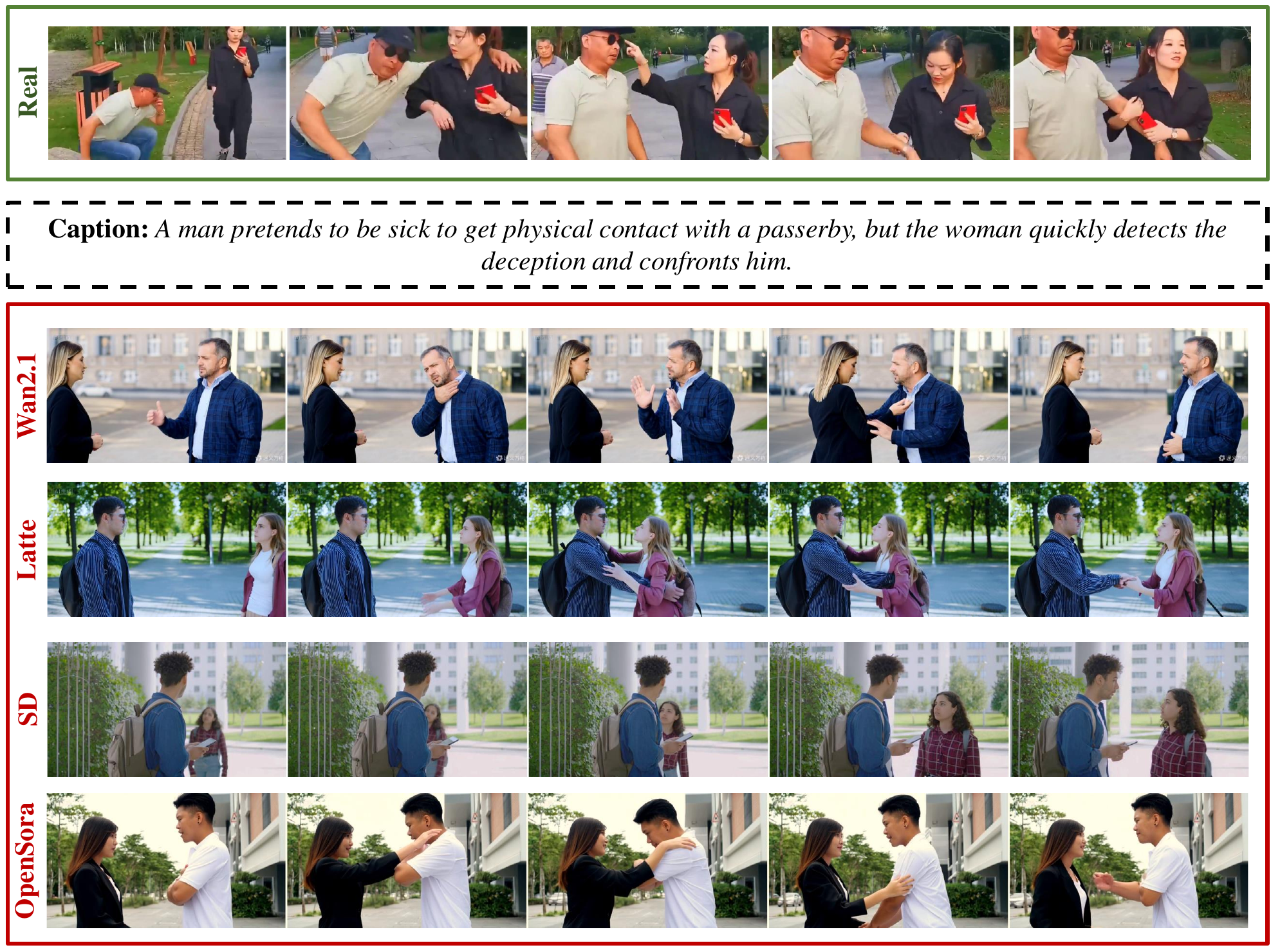}
	\caption{Example video samples from the SAVGE dataset. Each fake video is generated using the caption of a real video as the prompt, and the resulting real--fake set is organized as a semantic group.}
	\label{fig:showcase}
	\vspace{-1em}
\end{figure*}

To analyze the effect of the retrieval-based filtering strategy, we visualize the ``Fake Video--Caption'' and ``Fake Video--Real Video'' similarity-score distributions before and after the filtering process. As shown in Figure~\ref{fig:sim}, \rev{filtering shifts the similarity-score distributions toward higher values}, indicating that the filtering process removes generated samples with weak semantic correspondence to their paired real videos. This quantitative trend is consistent with the qualitative examples in Figure~\ref{fig:showcase}, where the retained fake videos generally preserve the primary subjects, scenes, and actions described by the captions of their corresponding real videos. Meanwhile, these examples still exhibit visible appearance discrepancies and generation artifacts, suggesting that filtering improves semantic consistency without removing the subtle cues required for detection. Overall, the proposed filtering strategy enhances semantic alignment while preserving the diversity and discriminative characteristics of fake samples needed for robust AIGV detection.

\subsection{Dataset Formulation}

\noindent Formally, SAVGE is organized as a series of semantic groups, with the $i$-th group denoted as:
\begin{align}
	G_i = \left\{V^{i,0}, \left\{V^{i,j}\right\}_{j=1}^{L_i}, T_i\right\}
\end{align}
where $V^{i,0}$ is the real video, $\{V^{i,j}\}_{j=1}^{L_i}$ is the set of fake videos generated from the same textual prompt $T_i$, and $L_i$ is the number of fake videos in the $i$-th group. For the training set, $4 \leq L_i \leq 6$, while for the test set, $L_i$ is fixed to $1$.

\rev{Table~\ref{tab:dataset} shows that the sources differ in resolution, frame rate, and duration. Semantic grouping matches content rather than these technical properties. Source and caption biases, VideoMAE-based selection, technical video differences, and the limited generator collection may therefore still affect detection.}

\vspace{-1em}
\vspace{-2em}

\section{EDSC: Explicit Disentangling and Semantic-aware Contrasting for AIGV Detection}

\noindent Building on the aligned structure of SAVGE, we introduce Explicit Disentangling and Semantic-aware Contrasting (EDSC), a training framework that moves AIGV detection beyond conventional binary supervision. Rather than treating semantics as an uncontrolled source of bias, EDSC uses the captions and semantic group annotations in SAVGE to disentangle semantic content from generation-related artifacts in the learned representation. This design is motivated by the observation that generalizable detectors should rely on forgery traces instead of category-, scene-, or action-level shortcuts, a principle also supported by recent generalizable deepfake detection research~\cite{ucf}. At a high level, EDSC uses textual annotations as semantic anchors to localize semantic factors in video features, suppresses these factors from the artifact branch, and further contrasts real and generated videos under matched semantic conditions. In this way, EDSC turns the semantic alignment of SAVGE into a form of supervision, enabling detectors to learn artifact-specific representations that capture more intrinsic differences between real and AI-generated videos.

\subsection{Overview}

\noindent EDSC is a model-agnostic training framework designed to exploit the semantic structure of SAVGE for robust AIGV detection. Given a video and its paired caption, EDSC uses a video encoder to produce two complementary visual representations: a visual semantic feature $\boldsymbol{f}_{SV}$, which is encouraged to capture the caption-related content of the video, and a visual artifact feature $\boldsymbol{f}_{AV}$, which is optimized to retain generation-specific cues for real--fake classification. In parallel, a text encoder maps the caption into a semantic reference feature $\boldsymbol{f}_{S}$, providing an explicit semantic anchor for aligning and isolating content information. During training, $\boldsymbol{f}_{SV}$ and $\boldsymbol{f}_{S}$ serve as auxiliary supervisory signals that guide disentanglement, while only $\boldsymbol{f}_{AV}$ is used as the final detection representation. This design encourages the detector to make decisions based on generation artifacts rather than semantic shortcuts such as objects, scenes, or actions.

\begin{figure*}[]
	\centering
	\includegraphics[width=\textwidth]{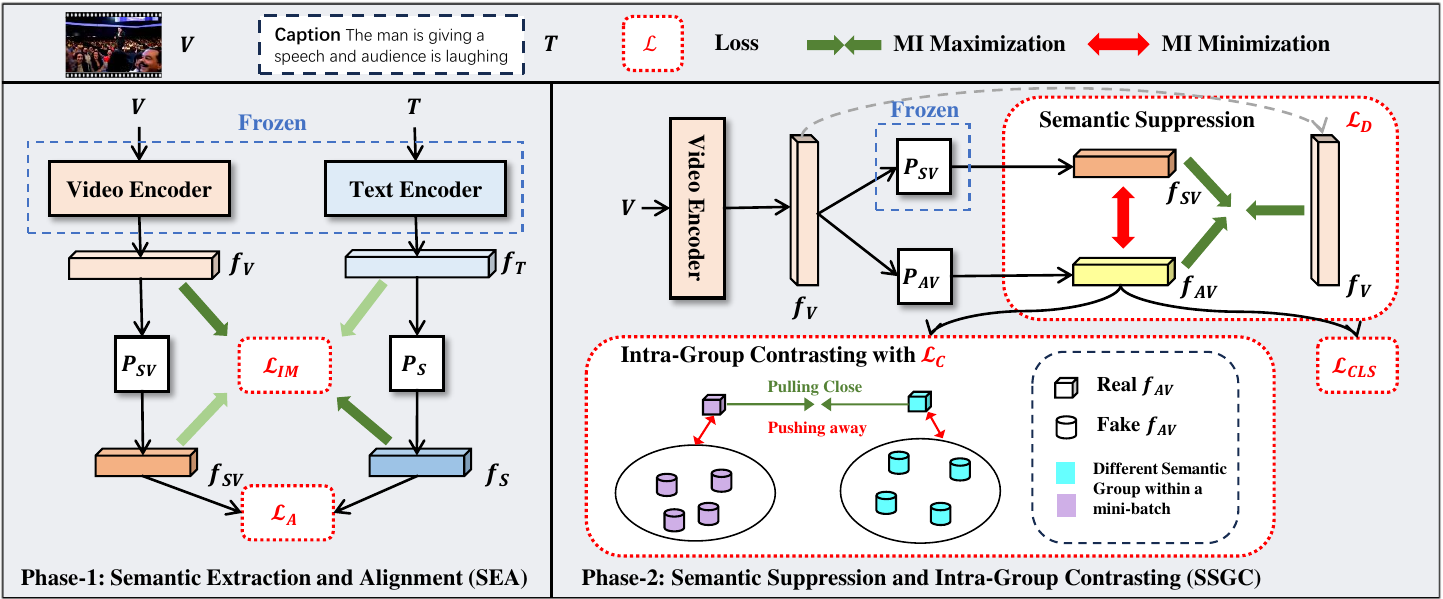}
	\caption{The overall framework of our proposed EDSC approach. It consists of two alternating phases: (1) Semantic Extraction and Alignment (SEA) and (2) Semantic Suppression and Intra-Group Contrasting (SSGC). \rev{The frozen boxes indicate fixed parameters in the corresponding phase, not fixed output features.}}
	\label{fig:framework}
	\vspace{-1em}
\end{figure*}

The overall pipeline of EDSC is illustrated in Figure~\ref{fig:framework}. EDSC is optimized through two alternating phases: Semantic Extraction and Alignment (SEA) and Semantic Suppression and Intra-Group Contrasting (SSGC). \rev{In SEA, EDSC maximizes cross-modal mutual information between each projected semantic feature and the basic feature from the other modality. It also minimizes the distance between $\boldsymbol{f}_{SV}$ and $\boldsymbol{f}_S$.} This objective encourages $\boldsymbol{f}_{SV}$ to encode caption-consistent content, such as objects, scenes, and actions, and therefore establishes an explicit semantic anchor inside the visual representation. By first localizing semantic information in a dedicated branch, EDSC provides a reliable target for removing content-related factors from the artifact branch in the following phase. In the SSGC phase, EDSC minimizes the intra-modal mutual information between $\boldsymbol{f}_{SV}$ and the visual artifact feature $\boldsymbol{f}_{AV}$, encouraging $\boldsymbol{f}_{AV}$ to discard semantic cues and preserve generation-related artifacts. Meanwhile, an intra-group contrastive objective is applied in the $\boldsymbol{f}_{AV}$ space, where real and fake videos sharing the same semantic description are contrasted to strengthen category separation under matched content conditions. The SEA and SSGC phases are alternated until convergence, yielding a semantically invariant artifact representation for robust real--fake classification. At inference time, EDSC requires only the video encoder and the artifact feature $\boldsymbol{f}_{AV}$, \rev{with no text input or semantic branch.}

\subsection{Encoders and Feature Extraction}

EDSC employs a video encoder and a text encoder to extract basic features from the two modalities. Notably, EDSC imposes no restrictions on the choice of encoders. For example, one may adopt a unified vision-language model such as XCLIP~\cite{XCLIP}, or alternatively combine two single-modal models such as VideoMAE~\cite{wang2023videomae} and BERT~\cite{BERT}. This flexibility makes EDSC adaptable to diverse encoder architectures and practical deployment requirements.

Specifically, we denote the basic features extracted from the video encoder as $\boldsymbol{f}_{V} \in \mathbb{R}^{d_V}$ and the features extracted from the text encoder as $\boldsymbol{f}_{T} \in \mathbb{R}^{d_T}$. We then project these features into a latent space using linear projection heads $P_{SV}$, $P_{AV}$, and $P_S$ to obtain the visual semantic feature $\boldsymbol{f}_{SV}$, visual artifact feature $\boldsymbol{f}_{AV}$, and semantic reference feature $\boldsymbol{f}_{S}$, respectively, as follows:
\begin{align}
\boldsymbol{f}_{SV} &= P_{SV}(\boldsymbol{f}_{V}),  \\
\boldsymbol{f}_{AV} &= P_{AV}(\boldsymbol{f}_{V}),\displaybreak[4]\\
\boldsymbol{f}_{S} &= P_{S}(\boldsymbol{f}_{T})
\end{align}
where $\boldsymbol{f}_{SV}$, $\boldsymbol{f}_{AV}$, $\boldsymbol{f}_{S} \in \mathbb{R}^d$ and $d$ is the dimension of the latent space. The three projected features are then used in the subsequent SEA and SSGC phases.

\subsection{Semantic Extraction and Alignment Phase}

\noindent The SEA phase aligns the visual semantic feature $\boldsymbol{f}_{SV}$ with the information described in the text annotation, preparing it to guide semantic suppression in the SSGC phase. To achieve this goal, 
EDSC 
 \rev{combines cross-modal mutual-information maximization with a distance constraint between the projected semantic features to encourage semantic alignment,} using Deep InfoMax~\cite{DIM} with Jensen-Shannon divergence (JSD-DIM)~\cite{JSD}, which provides stable mutual information estimation.
The JSD-DIM estimator ${I}_{\theta}^{(J)}(X,Y)$ for two generic random variables $X \in \mathcal{X}$ and $Y \in \mathcal{Y}$ is formulated as follows:
\begin{equation}
\begin{aligned}
    {I}_{\theta}^{(J)}\left(X,Y\right) &= \mathbb{E}_{p(x,y)}\left[-\log\left(1+e^{-T_\theta(x,y)}\right)\right] \\
    &- \mathbb{E}_{p(x)p(y)}\left[\log\left(1+e^{T_\theta(x,y)}\right)\right]
\end{aligned}
\end{equation}
where $T_\theta: \mathcal{X} \times \mathcal{Y} \mapsto \mathbb{R}$ is a learnable statistics network parameterized by $\theta$. Maximizing ${I}_{\theta}^{(J)}(X,Y)$ estimates and increases the mutual information between $X$ and $Y$.

During this phase, we freeze the \rev{video and text encoder parameters} and optimize the semantic projection heads $P_{SV}$ and $P_S$. The phase objective $\mathcal{L}_{SEA}$ is formulated as follows:
\begin{align}
    \mathcal{L}_{IM} &= {I}_{\theta_{VT}}^{(J)}\left(\boldsymbol{f}_{SV}, \boldsymbol{f}_{T}\right) +{I}_{\theta_{TV}}^{(J)}\left(\boldsymbol{f}_{S}, \boldsymbol{f}_{V}\right), \\
    \mathcal{L}_{A} &= \lVert\boldsymbol{f}_{SV} - \boldsymbol{f}_{S}\rVert_2, \\
    \mathcal{L}_{SEA} &= -\mathcal{L}_{IM} + \mathcal{L}_{A}
\end{align}
$\mathcal{L}_{SEA}$ comprises two terms. The first term, $\mathcal{L}_{IM}$, maximizes the cross-modal mutual information between the projected semantic features $\boldsymbol{f}_{SV}$/$\boldsymbol{f}_{S}$ and the basic features $\boldsymbol{f}_{T}$/$\boldsymbol{f}_{V}$ from the other modality. Since the encoders are frozen in this phase, $\mathcal{L}_{IM}$ encourages $\boldsymbol{f}_{SV}$ and $\boldsymbol{f}_{S}$ to capture the semantic information shared between the two modalities. The second term, $\mathcal{L}_{A}$, minimizes the distance between $\boldsymbol{f}_{SV}$ and $\boldsymbol{f}_{S}$, further enforcing semantic alignment. Overall, $\mathcal{L}_{SEA}$ encourages $\boldsymbol{f}_{SV}$ to learn video semantic components that are aligned with the information described in the text annotation.

\subsection{Semantic Suppression and Intra-Group Contrasting  Phase}

\noindent The SSGC phase aims to transform the artifact branch into a discriminative yet semantically invariant representation. It contains two complementary objectives. First, using the aligned semantic feature $\boldsymbol{f}_{SV}$ learned in the SEA phase as guidance, SSGC suppresses the semantic information contained in the visual artifact feature $\boldsymbol{f}_{AV}$ while preserving the non-semantic information derived from the basic video feature $\boldsymbol{f}_{V}$. This encourages $\boldsymbol{f}_{AV}$ to focus on generation-related artifacts rather than caption-level content. Second, SSGC applies intra-group contrastive learning in the $\boldsymbol{f}_{AV}$ space, where real and fake videos with the same semantic description are explicitly separated. By combining semantic suppression with contrastive separation under aligned content, SSGC produces an artifact-focused feature $\boldsymbol{f}_{AV}$ that is more suitable for robust real--fake classification.

To achieve feature disentanglement, we adopt JSD-DIM for intra-modal mutual information estimation as follows:
\begin{align}
    \mathcal{L}_{D} &= -{I}_{\theta_{AS}}^{(J)}\left(\boldsymbol{f}_{AV}, \boldsymbol{f}_{SV}\right) 
    + {I}_{\theta_{W}}^{(J)}\left([\boldsymbol{f}_{SV}||\boldsymbol{f}_{AV}], \boldsymbol{f}_{V}\right) 
\end{align}
By minimizing ${I}_{\theta_{AS}}^{(J)}\left(\boldsymbol{f}_{AV}, \boldsymbol{f}_{SV}\right)$, we encourage $\boldsymbol{f}_{AV}$ to capture information that is distinct from $\boldsymbol{f}_{SV}$. However, this term alone may lead to trivial solutions. To mitigate this issue, the second term ${I}_{\theta_{W}}^{(J)}\left([\boldsymbol{f}_{SV}||\boldsymbol{f}_{AV}], \boldsymbol{f}_{V}\right)$ encourages $\boldsymbol{f}_{AV}$ to retain information that is complementary to $\boldsymbol{f}_{SV}$ and sufficient to recover the information contained in the basic feature $\boldsymbol{f}_V$. These two terms jointly produce an explicitly disentangled $\boldsymbol{f}_{AV}$ with reduced semantic information.
We maximize $\mathcal{L}_{D}$ (equivalently, minimize $-\mathcal{L}_{D}$ in $\mathcal{L}_{SSGC}$), thereby minimizing ${I}_{\theta_{AS}}^{(J)}$ while maximizing ${I}_{\theta_{W}}^{(J)}$.

Furthermore, we apply semantic group-level feature contrasting to enforce a clearer separation between the two categories. To this end, we construct a mini-batch of $N=(1+L)M$ samples, including $M$ real--fake video groups. The $i$-th group contains one real video and $L$ fake videos sharing the same textual annotation, whose visual artifact features are denoted as $\boldsymbol{f}_{AV}^{i,0}$ and $\{\boldsymbol{f}_{AV}^{i,k}\}_{k=1}^L$, respectively. We use the InfoNCE loss~\cite{infonce} to formulate the learning objective:
\begin{align}
    \mathcal{L}_{C} &= -\frac{1}{M}\sum_{i=1}^M \log \frac{sim(\boldsymbol{f}_{AV}^{i,0}, \boldsymbol{f}_{AV}^{i,*})}{\sum_{k\in[L]\cup\{*\}} sim(\boldsymbol{f}_{AV}^{i,0}, \boldsymbol{f}_{AV}^{i,k})}
\end{align}
where $[L]=\{1,\ldots,L\}$, $\boldsymbol{f}_{AV}^{i,*}$ is sampled from all other real features $\{\boldsymbol{f}_{AV}^{j,0}\}_{j\neq i}$ within the mini-batch as a positive sample, and $sim(f,g)=\exp(f\cdot g/\tau)$ is the similarity function. Intuitively, $\mathcal{L}_{C}$ pulls the real visual artifact feature $\boldsymbol{f}_{AV}^{i,0}$ close to the positive sample $\boldsymbol{f}_{AV}^{i,*}$ while pushing it away from all fake features $\{\boldsymbol{f}_{AV}^{i,k}\}_{k=1}^L$ within the same semantic group. 

Finally, $\boldsymbol{f}_{AV}$ is used for classification. 
We apply a linear transformation followed by a sigmoid function to map each $\boldsymbol{f}_{AV}^{i, k}$ to a probability score $\hat{y}^{i, k}$. The classification loss is formulated as a weighted binary cross-entropy loss:
Real videos are assigned label $1$ and fake videos label $0$; the real-video term is weighted by $L$ to balance the $L$ fake samples in each group.
\begin{align}
    \mathcal{L}_{CLS} &= -\frac{1}{M}\sum_{i=1}^M \left[L\log \hat{y}^{i,0} + \sum_{k=1}^L\log(1-\hat{y}^{i,k})\right]
\end{align}
The overall loss function for the SSGC phase is formulated as:
\begin{align}
    \mathcal{L}_{SSGC} &= -\mathcal{L}_{D} + \mathcal{L}_{C} + \mathcal{L}_{CLS}
\end{align}

\subsection{Training and Inference Scheme}

\noindent Algorithm~\ref{alg:edsc} summarizes the training and inference pipeline of EDSC. In particular, it describes the alternating optimization scheme adopted during training, where the model first performs the SEA phase to align the visual semantic representation with the textual semantic anchor, and then executes the SSGC phase to suppress semantic information in the artifact branch while enforcing intra-group discrimination between real and fake samples. This alternating design enables the semantic branch to provide \rev{guidance from the learned semantic projection} for disentanglement while progressively refining the semantically invariant artifact representation used for final prediction. During inference, only the visual encoder $E_V$, the projection head $P_{AV}$, and the classifier $C$ are retained\rev{; neither captions nor the semantic branch are needed.} \rev{In SSGC, $E_V$, $P_{AV}$, and $C$ are updated; $P_{SV}$ is frozen, and the text branch remains unchanged. The output of $P_{SV}$ may still change as $E_V$ is updated.}
\begin{algorithm}[H]
	\small
	\caption{Training and Inference of EDSC}
	\label{alg:edsc}
	\begin{algorithmic}[1]
		\Require Video encoder $E_V$; text encoder $E_T$; projection heads $P_{SV}$, $P_{AV}$, and $P_S$; classifier $C$; SAVGE training set with semantic groups; number of training iterations $T$.
		\Ensure Trained video encoder $E_V$, projection head $P_{AV}$, and classifier $C$.
		\State \textbf{Training:} Initialize the parameters of $E_V$, $E_T$, $P_{SV}$, $P_{AV}$, $P_S$, and $C$.
		\For{$t=1$ to $T$}
		\State Sample a mini-batch of $M$ semantic groups, each containing one real video, $L$ fake videos, and the shared text annotation.
		\State Extract basic video features $\boldsymbol{f}_V \leftarrow E_V(\boldsymbol{x})$ and text features $\boldsymbol{f}_T \leftarrow E_T(\boldsymbol{s})$.
		\State Obtain $\boldsymbol{f}_{SV} \leftarrow P_{SV}(\boldsymbol{f}_V)$, $\boldsymbol{f}_{AV} \leftarrow P_{AV}(\boldsymbol{f}_V)$, and semantic reference feature $\boldsymbol{f}_S \leftarrow P_S(\boldsymbol{f}_T)$.
		\If{$t$ is in the SEA phase}
		\State \rev{Compute $\mathcal{L}_{SEA}=-\mathcal{L}_{IM}+\mathcal{L}_A$.}
		\State \rev{Update $P_{SV}$ and $P_S$ using $\mathcal{L}_{SEA}$; freeze $E_V$ and $E_T$.}
		\Else
		\State Compute disentanglement loss $\mathcal{L}_{D}$ between $\boldsymbol{f}_{SV}$, $\boldsymbol{f}_{AV}$, and $\boldsymbol{f}_V$.
		\State Compute intra-group contrastive loss $\mathcal{L}_{C}$ on $\boldsymbol{f}_{AV}$ within each semantic group.
		\State Compute classification loss $\mathcal{L}_{CLS}$ from $\hat{y}=C(\boldsymbol{f}_{AV})$.
		\State Form $\mathcal{L}_{SSGC} \leftarrow -\mathcal{L}_{D} + \mathcal{L}_{C} + \mathcal{L}_{CLS}$.
		\State Update $E_V$, $P_{AV}$, and $C$ using $\mathcal{L}_{SSGC}$; \rev{freeze $P_{SV}$.}
		\EndIf
		\EndFor
		\State \textbf{Inference:} For a test video, compute $\boldsymbol{f}_V \leftarrow E_V(\boldsymbol{x})$, $\boldsymbol{f}_{AV} \leftarrow P_{AV}(\boldsymbol{f}_V)$, and output the prediction $\hat{y}=C(\boldsymbol{f}_{AV})$.
	\end{algorithmic}
\end{algorithm}

\section{Experiments}\label{sec:exp}

\subsection{Datasets}

\noindent We conduct a comprehensive evaluation of EDSC on the SAVGE dataset, which is divided into a development set and a test set for training and benchmarking, respectively. Detailed statistics of the SAVGE development and test sets are provided in Table~\ref{tab:dataset}. For model development, we further split the development set into training and validation subsets with a ratio of 9:1.
In addition, we perform cross-dataset evaluation on GenVideo~\cite{DeMamba} to assess the generalization capability of EDSC and compare the relative difficulty of the two benchmarks.

\subsection{Comparative Methods}

\noindent Since EDSC is model-agnostic, it can be applied to a wide range of AIGV detection methods. We select several state-of-the-art methods for comparison, including three frame-level methods, F$^3$Net~\cite{f3net}, NPR~\cite{NPR}, and CLIP-B-FT~\cite{CLIP}, and three video-level methods, TALL~\cite{TALL}, XCLIP-B-FT~\cite{XCLIP}, and DeMamba-XCLIP-B-FT~\cite{DeMamba}. For methods using CLIP or XCLIP, we adopt their respective text encoders to extract the semantic reference feature $\boldsymbol{f}_{S}$; for the other methods, we use a pre-trained BERT-base model~\cite{BERT}.

\subsection{Implementation Details}

\begin{table*}[b]
	\vspace{-1em}
		\caption{In-dataset and cross-dataset evaluation on the SAVGE and GenVideo~\cite{DeMamba} datasets. All methods below are trained solely on the SAVGE development set.}\label{tab:main}
		\setlength{\tabcolsep}{7pt}
		\begin{tabular}{cc|c|ccc||ccc}
			\toprule
			\multirow{2}{*}{\textbf{Method}} & {\textbf{Detection}}& \multirow{2}{*}{\textbf{EDSC}} & \multicolumn{3}{c||}{\textbf{Tested on SAVGE}} &\multicolumn{3}{c}{\textbf{Tested on GenVideo}}\\\cmidrule(lr){4-9}
			& {\textbf{Level}} & & \textbf{Recall} & \textbf{F1} & \textbf{AP}  & \textbf{Recall} & \textbf{F1} & \textbf{AP}\\\midrule
			\multirow{2}{*}{BERT+F$^3$Net~\cite{f3net}} & \multirow{2}{*}{Frame} & - & 0.7371 & 0.7634 & 0.8123 & 0.8032 & 0.7961 & 0.8655\\
			& & \checkmark & 0.7962 & 0.7825 & 0.8569 & 0.8683 & 0.8912 & 0.9306\\\hline
			\multirow{2}{*}{BERT+NPR~\cite{NPR}} & \multirow{2}{*}{Frame} & - & 0.7832 & 0.7213 & 0.7944 & 0.8235 & 0.8162 & 0.7937\\
			& & \checkmark & 0.8345 & 0.7921 & 0.8212 & 0.8912 & 0.8657 & 0.8345\\\hline
			\multirow{2}{*}{CLIP-B-FT~\cite{CLIP}} & \multirow{2}{*}{Frame} & - & 0.8341 & 0.8235 & 0.8654 & 0.9291 & 0.8065 & 0.9344\\
			& & \checkmark & 0.8587 & 0.8452 & 0.8341 & {0.9344} & 0.9127 & 0.9215 \\\hline
			\multirow{2}{*}{BERT+TALL~\cite{TALL}} & \multirow{2}{*}{Video} & - & 0.7663 & 0.7219 & 0.7891 & 0.8396 & 0.8215 & 0.8568\\
			& & \checkmark & 0.7978 & 0.7452 & 0.8345 & 0.8412 & 0.8097 & 0.8701\\\hline
			\multirow{2}{*}{XCLIP-B-FT~\cite{XCLIP}} & \multirow{2}{*}{Video} & - & 0.7954 & 0.7059 & 0.8096 & 0.8471 & 0.8042 & 0.8760\\
			& & \checkmark & 0.8513 & 0.8122 & 0.8520 & 0.9295 & 0.8650 & 0.9487\\\hline
			\multirow{2}{*}{DeMamba-XCLIP-B-FT~\cite{DeMamba}} & \multirow{2}{*}{Video} & - & 0.8568 & 0.8476 & 0.8719 & 0.9457 & \textbf{0.9143} & 0.9638\\
			& & \checkmark & \textbf{0.8712} & \textbf{0.8651} & \textbf{0.8760} & \textbf{0.9503} & 0.9029 & \textbf{0.9794}\\\bottomrule
		\end{tabular}
\end{table*}

\noindent For methods without EDSC, we adopt a unified training protocol with a batch size of 128, a learning rate of $1\times 10^{-6}$, and the AdamW optimizer for 10 epochs. Each mini-batch is constructed by sampling 64 real videos and 64 fake videos to maintain class balance during training.

For methods with EDSC, we construct each mini-batch from 25 semantic groups, with each group containing 1 real video and 4 fake videos, resulting in a batch size of 125. We use the same learning rate of $1\times 10^{-6}$ and the AdamW optimizer, and train for 10 epochs. In the first 2,000 iterations of each epoch, we optimize only the SEA phase and then switch to the SSGC phase for the remaining iterations. The output dimensions of the projection heads $P_{SV}$, $P_{AV}$, and $P_S$ are set to $d=512$. The structure of the statistics network $T_\theta$ follows the original Deep InfoMax design~\cite{DIM}. The temperature $\tau$ in $\mathcal{L}_{C}$ is set to 0.1 for all experiments.

\rev{Each backbone uses the same frame sampling and input size with and without EDSC.}

For evaluation metrics, we report average precision (AP), recall, and the F1 score to assess detector effectiveness. Recall and the F1 score are computed using a decision threshold of 0.5, whereas AP is threshold-independent. 
For frame-based detection methods, we average frame-level predictions to obtain the corresponding video-level predictions. All results are reported as the average of five independent runs.

\subsection{In-dataset Evaluation on the SAVGE Dataset}

\noindent We first evaluate a diverse set of baseline methods on the proposed SAVGE dataset, with the results reported in Table~\ref{tab:main}. Among all methods without EDSC, DeMamba-XCLIP-B-FT~\cite{DeMamba} achieves the strongest overall performance, with an F1 score of 0.8476 and an AP of 0.8719, outperforming both frame-level and video-level counterparts. This advantage is likely attributable to its dedicated Mamba-based temporal modeling module, which is well suited to capturing complex spatiotemporal dependencies in video content. The strong baseline performance of DeMamba-XCLIP-B-FT also indicates the importance of effective temporal modeling for AIGV detection on semantically aligned video benchmarks.

After incorporating EDSC, all methods improve on most evaluation metrics, demonstrating that the proposed framework \rev{can improve most metrics for detectors} built on different backbone architectures. In particular, DeMamba-XCLIP-B-FT~\cite{DeMamba}+EDSC achieves the best overall performance among all compared methods, reaching an F1 score of 0.8651 and an AP of 0.8760. These gains suggest that EDSC complements strong baseline encoders by further suppressing semantic interference and promoting artifact-sensitive representations, thereby improving discriminability even when the underlying detector is already highly competitive.

XCLIP-B-FT~\cite{XCLIP} \rev{shows the largest F1 gain from EDSC}, with improvements of approximately \rev{0.06}/0.11/0.04 in recall/F1/AP, respectively. One possible explanation is that XCLIP, as a unified vision-language model, is naturally compatible with the semantic supervision introduced by EDSC and can therefore exploit the disentanglement objective more effectively. Taken together, these results show that EDSC is a general and effective training framework for improving the accuracy and robustness of AIGV detectors on SAVGE.

\begin{table*}[b]
 \vspace{-1em}  
    \centering
    \caption{Robustness evaluation on the SAVGE dataset by applying various data transformations.}\label{robustness_results}
    \setlength{\tabcolsep}{4pt}
    \begin{tabular}{c|c|c|ccccc}
    \toprule
    \textbf{Method} & \textbf{Metric} & \textbf{Original} & \textbf{JPEG}  & \textbf{Gaussian} & \textbf{Flip} & \textbf{Crop} & \textbf{Color Trans.}\\
\midrule
\multirow{3}{*}{BERT+TALL~\cite{TALL}} & \textbf{Recall} & 0.7663 & ${0.7359}_{\cl{-0.03}}$ & ${0.7548}_{\cl{-0.01}}$ & ${0.7515}_{\cl{-0.01}}$ & ${0.6029}_{\cl{-0.16}}$ & ${0.7350}_{\cl{-0.03}}$ \\
        & \textbf{F1} & 0.7219 & ${0.5483}_{\cl{-0.17}}$ & ${0.6735}_{\cl{\rev{-0.05}}}$ & ${0.7003}_{\cl{-0.02}}$ & ${0.4625}_{\cl{\rev{-0.26}}}$ & ${0.6968}_{\cl{\rev{-0.03}}}$ \\
        & \textbf{AP} & 0.7891 & ${0.6673}_{\cl{-0.12}}$ & ${0.6856}_{\cl{-0.10}}$ & ${0.7596}_{\cl{\rev{-0.03}}}$ & ${0.4715}_{\cl{\rev{-0.32}}}$ & ${0.7616}_{\cl{\rev{-0.03}}}$\\
\midrule
BERT+TALL~\cite{TALL} & \textbf{Recall} & 0.7978 & ${0.7661}_{\cl{-0.03}}$ & ${0.7850}_{\cl{-0.01}}$ & ${0.7947}_{\cl{-0.00}}$ & ${0.6292}_{\cl{-0.17}}$ & ${0.7939}_{\cl{-0.00}}$\\
  +      & \textbf{F1} & 0.7452 & ${0.6415}_{\cl{-0.10}}$ & ${0.6892}_{\cl{-0.06}}$ & ${0.7327}_{\cl{-0.01}}$ & ${0.5895}_{\cl{-0.16}}$ & ${0.7439}_{\cl{-0.00}}$\\
   EDSC     & \textbf{AP} & 0.8345 & ${0.7762}_{\cl{-0.06}}$ & ${0.7765}_{\cl{-0.06}}$ & ${0.8081}_{\cl{-0.03}}$ & ${0.5794}_{\cl{-0.26}}$ & ${0.8343}_{\cl{-0.00}}$
       \\        
\midrule    

\multirow{3}{*}{XCLIP-B-FT~\cite{XCLIP}} & \textbf{Recall} & 0.7954 & ${0.7646}_{\cl{-0.03}}$ & ${0.7813}_{\cl{-0.01}}$ & ${0.7805}_{\cl{-0.01}}$ & ${0.6946}_{\cl{-0.10}}$ & ${0.7732}_{\cl{-0.02}}$  \\
        & \textbf{F1} & 0.7059 & ${0.5758}_{\cl{-0.13}}$ & ${0.6736}_{\cl{-0.03}}$ & ${0.6771}_{\cl{-0.03}}$ & ${0.4551}_{\cl{-0.25}}$ & ${0.6730}_{\cl{-0.03}}$ \\
        & \textbf{AP} & 0.8096 & ${0.6286}_{\cl{-0.18}}$ & ${0.7250}_{\cl{-0.08}}$ & ${0.7882}_{\cl{-0.02}}$ & ${0.4987}_{\cl{-0.31}}$ & ${0.8002}_{\cl{-0.01}}$ \\
\midrule
XCLIP-B-FT~\cite{XCLIP} & \textbf{Recall} & 0.8513 & ${0.8097}_{\cl{-0.04}}$ & ${0.8204}_{\cl{-0.03}}$ & ${0.8339}_{\cl{-0.02}}$ & ${0.6963}_{\cl{-0.16}}$ & ${0.8474}_{\cl{-0.00}}$ \\
  +      & \textbf{F1} & 0.8122 & ${0.7194}_{\cl{-0.09}}$ & ${0.7646}_{\cl{-0.05}}$ & ${0.7926}_{\cl{-0.02}}$ & ${0.7069}_{\cl{-0.11}}$ & ${0.8051}_{\cl{-0.01}}$\\
   EDSC     & \textbf{AP} & 0.8520 & ${0.7539}_{\cl{-0.10}}$ & ${0.8016}_{\cl{-0.05}}$ & ${0.8338}_{\cl{-0.02}}$ & ${0.5750}_{\cl{-0.28}}$ & ${0.8415}_{\cl{-0.01}}$ 
       \\        
\midrule

\multirow{3}{*}{DeMamba-XCLIP-B-FT~\cite{DeMamba}} & \textbf{Recall} & 0.8568 & $0.8371_{\cl{-0.02}}$ & $0.8419_{\cl{-0.01}}$ & $0.8542_{\cl{-0.00}}$ & $0.6794_{\cl{-0.18}}$ & $0.8514_{\cl{-0.01}}$ \\
        & \textbf{F1} & 0.8476 & $0.7043_{\cl{-0.14}}$ & $0.7841_{\cl{-0.06}}$ & $0.8399_{\cl{-0.01}}$ & $0.6350_{\cl{-0.21}}$ & $0.8471_{\cl{-0.00}}$ \\
        & \textbf{AP} & 0.8719 & $0.6590_{\cl{-0.21}}$ & $0.7764_{\cl{-0.10}}$ & $0.8711_{\cl{-0.00}}$ & $0.6039_{\cl{-0.27}}$ & $0.8698_{\cl{-0.00}}$\\
\midrule
DeMamba-XCLIP-B-FT~\cite{DeMamba} & \textbf{Recall} & {0.8712} & ${0.8643}_{\cl{-0.01}}$ & ${0.8702}_{\cl{-0.00}}$ & $0.8710_{\cl{-0.00}}$ & $0.7095_{\cl{-0.16}}$&$0.8686_{\cl{-0.00}}$\\
  +      & \textbf{F1} & 0.8651 & $0.7597_{\cl{\rev{-0.11}}}$ & $0.8169_{\cl{\rev{-0.05}}}$ & $0.8633_{\cl{-0.00}}$ &$0.6398_{\cl{-0.23}}$&$0.8624_{\cl{-0.00}}$\\
   EDSC     & \textbf{AP} & 0.8760 & $0.7410_{\cl{-0.14}}$ & $0.8281_{\cl{-0.05}}$
        & $0.8703_{\cl{-0.01}}$ & $0.6184_{\cl{-0.26}}$&$0.8755_{\cl{-0.00}}$
       \\
        \bottomrule
    \end{tabular}

\end{table*}

\subsection{Cross-dataset Evaluation on the GenVideo Dataset}

\noindent We further evaluate the generalization ability of EDSC on the GenVideo dataset~\cite{DeMamba} by directly testing models trained on the SAVGE development set on the GenVideo test set. This setting provides a more stringent assessment of cross-dataset transferability, since the training and evaluation data differ in collection protocol, semantic organization, and sample construction.
As reported in Table~\ref{tab:main}, methods equipped with EDSC improve most backbone--metric combinations relative to their corresponding baseline versions, demonstrating that the representations learned by EDSC generalize beyond the SAVGE benchmark. Among all compared methods, DeMamba-XCLIP-B-FT~\cite{DeMamba}+EDSC \rev{achieves the highest cross-dataset recall and AP, at 0.9503 and 0.9794, respectively, although its F1 decreases.} These results further suggest that the gains brought by EDSC are not limited to in-dataset fitting, but extend to scenarios in which detectors are applied to previously unseen data distributions.

\rev{Most reported metrics are higher on GenVideo than on SAVGE. The datasets differ in their real-video sources, generation pipelines, and filtering, so these results do not identify a single cause for that difference. SAVGE specifically supports evaluation under caption-based semantic grouping; the cross-dataset results test whether the learned representations also transfer to another collection.}

\rev{EDSC improves 32 of the 36 means in Table~\ref{tab:main}. CLIP-B-FT AP decreases by 3.13 and 1.29 percentage points on SAVGE and GenVideo, respectively, suggesting that semantic suppression may also remove useful ranking information. On GenVideo, BERT+TALL and DeMamba F1 decrease by 1.18 and 1.14 points despite higher AP: better ranking need not improve F1 at the fixed 0.5 threshold. These five-run means show gains across most settings, but do not establish the significance of small differences.}

\subsection{Robustness Evaluation on the SAVGE Dataset}

\noindent Following the evaluation protocol in~\cite{DeMamba}, we assess robustness on the SAVGE dataset by applying several common data transformations to the original videos, including JPEG compression, Gaussian noise, frame flipping, cropping, and color transformation. Under this setting, we evaluate BERT+TALL~\cite{TALL}, XCLIP-B-FT~\cite{XCLIP}, and DeMamba-XCLIP-B-FT~\cite{DeMamba}, together with their EDSC-enhanced counterparts. The corresponding results are reported in Table~\ref{robustness_results}. \rev{We use frame-wise JPEG quality 35 and additive zero-mean Gaussian noise with $\sigma=0.05$ on $[0,1]$ pixels. Crop width and height are each 160--208 pixels on $224\times224$ frames; the same window is used across a video and resized back to $224\times224$.}
All model variants largely preserve their performance under frame flipping and color transformation. This behavior is expected, since these transformations neither substantially alter semantic content nor introduce pronounced artifacts that interfere with detection.
In contrast, cropping leads to severe performance degradation for DeMamba-XCLIP-B-FT and its EDSC-enhanced counterpart, with AP drops of 0.27 and 0.26, respectively. A plausible explanation is that cropping may remove visually informative regions that contain crucial generation artifacts, preventing the detectors from forming reliable predictions from incomplete spatial evidence.
JPEG compression and Gaussian noise also reduce detection performance, although their effects differ. In particular, DeMamba-XCLIP-B-FT~\cite{DeMamba} suffers a substantial AP drop of 0.21 after JPEG compression, highlighting video compression as a major challenge for robust AIGV detection. Although DeMamba-XCLIP-B-FT~\cite{DeMamba}+EDSC also degrades under JPEG compression, the AP drop is reduced to 0.14, indicating that EDSC improves resilience to compression artifacts to some extent. By comparison, the impact of Gaussian noise is more moderate, with AP drops of 0.10 and 0.05 for DeMamba-XCLIP-B-FT~\cite{DeMamba} and DeMamba-XCLIP-B-FT~\cite{DeMamba}+EDSC, respectively. Overall, these results suggest that, while EDSC remains vulnerable to severe degradations such as cropping and JPEG compression, it learns more robust representations under milder perturbations such as Gaussian noise, highlighting its potential for deployment in challenging real-world scenarios.

\begin{table*}[t]
    
    \caption{Ablation study on the effectiveness of various EDSC designs. The methods are trained and tested on the SAVGE dataset.}\label{tab:ablation}
    \centering
    \setlength{\tabcolsep}{5.5pt}
    \begin{tabular}{c|ccc|ccc|ccc}
        \toprule
        \multirow{2}{*}{\textbf{Method}} & \multicolumn{3}{c|}{\textbf{BERT+TALL}} & \multicolumn{3}{c}{\textbf{XCLIP-B-FT}} &\multicolumn{3}{|c}{\textbf{DeMamba-XCLIP-B-FT}}\\\cmidrule(lr){2-10}
        & \textbf{Recall} & \textbf{F1} & \textbf{AP} & \textbf{Recall} & \textbf{F1} & \textbf{AP}  & \textbf{Recall} & \textbf{F1} & \textbf{AP}\\
        \midrule
        w/o EDSC&0.7663&0.7219&0.7891 & 0.7954 & 0.7059 & 0.8096&0.8568&0.8476&0.8719 \\
        {EDSC} &0.7978&0.7452&0.8345& 0.8513 & 0.8122 & 0.8520 &0.8712&0.8651&0.8760\\\midrule
        \multicolumn{10}{c}{Ablation study on SEA Phase}\\
        \midrule
        {SEA w/o $\mathcal{L}_{IM}$}&0.7058&0.6154&0.6927 & 0.6741 & 0.6239 & 0.7116 &0.7336&0.7160&0.7397\\
        {SEA w/o $\mathcal{L}_A$} &0.7854&0.7148&0.8128& 0.8453 & 0.7972 & 0.8461 &0.8644&0.8279&0.8497\\
        \midrule
        \multicolumn{10}{c}{Ablation study on SSGC Phase}\\
        \midrule
        {SSGC w/o $\mathcal{L}_D$}&0.7601&0.7034&0.7798 & 0.8320 & 0.7761 & 0.8128 &0.8315&0.8142&0.8496\\
        {SSGC w/o $\mathcal{L}_C$} &0.7765&0.7107&0.7956& 0.8427 & 0.7856 & 0.8267 &0.8615&0.8342&0.8548\\\midrule
        \multicolumn{10}{c}{Ablation study on training strategy}\\
        \midrule
        EDSC w/ Joint Train &0.4767&0.4133&0.3759& 0.5395 & 0.4678 & 0.3916&0.5985&0.5037&0.4668 \\\bottomrule
    \end{tabular}
   
\end{table*}

\subsection{Ablation Study}

\noindent We conduct ablation studies to examine the contribution of different components in EDSC. We adopt BERT+TALL~\cite{TALL}, XCLIP-B-FT~\cite{XCLIP}, and DeMamba-XCLIP-B-FT~\cite{DeMamba} as backbone models and evaluate the following variants:

\subsubsection{SEA Phase} 
\noindent We first evaluate the effectiveness of the two objectives used in the SEA phase by separately removing the mutual information term $\mathcal{L}_{IM}$ and the feature alignment term $\mathcal{L}_{A}$. As shown in Table~\ref{tab:ablation}, removing $\mathcal{L}_{IM}$ causes a substantial performance drop across all three backbones. For example, on XCLIP-B-FT, recall/F1/AP decrease from 0.8513/0.8122/0.8520 with the full EDSC framework to 0.6741/0.6239/0.7116 without $\mathcal{L}_{IM}$. Similar degradations are observed on BERT+TALL and DeMamba-XCLIP-B-FT. This result indicates that cross-modal mutual information maximization is essential for extracting reliable semantic factors from the video representation because it explicitly encourages the visual semantic branch to preserve information shared with the textual caption. Without this term, the semantic branch cannot provide a stable anchor for subsequent semantic suppression in the artifact branch.

Removing $\mathcal{L}_{A}$ leads to a milder but still consistent degradation compared with the full EDSC framework. In particular, XCLIP-B-FT achieves 0.8453/0.7972/0.8461 in recall/F1/AP without $\mathcal{L}_{A}$, which is lower than the full EDSC result of 0.8513/0.8122/0.8520. This suggests that, although mutual information alignment captures shared cross-modal semantics, the direct distance constraint between $\boldsymbol{f}_{SV}$ and $\boldsymbol{f}_{S}$ further improves the compactness and consistency of the semantic space. Overall, these results demonstrate that $\mathcal{L}_{IM}$ plays the dominant role in semantic extraction, while $\mathcal{L}_{A}$ provides complementary alignment that stabilizes the SEA phase and improves the final artifact representation.

\subsubsection{Feature Disentanglement in SSGC Phase} 
\noindent We next examine the effect of feature disentanglement by removing the disentanglement loss $\mathcal{L}_{D}$ from the SSGC phase. In this variant, $\boldsymbol{f}_{AV}$ is no longer explicitly constrained to be independent of the semantic feature $\boldsymbol{f}_{SV}$, and the detector is optimized mainly by the contrastive loss $\mathcal{L}_{C}$ and classification loss $\mathcal{L}_{CLS}$. As shown in Table~\ref{tab:ablation}, removing $\mathcal{L}_{D}$ consistently degrades performance across the three backbones. For XCLIP-B-FT, recall/F1/AP drop from 0.8513/0.8122/0.8520 with the full EDSC framework to 0.8320/0.7761/0.8128. A similar decline is observed for DeMamba-XCLIP-B-FT, whose recall/F1/AP decrease from 0.8712/0.8651/0.8760 to 0.8315/0.8142/0.8496. These results indicate that explicit semantic suppression is crucial for preventing the artifact branch from retaining content-related shortcuts. By reducing the dependence between $\boldsymbol{f}_{AV}$ and $\boldsymbol{f}_{SV}$, $\mathcal{L}_{D}$ guides the detector toward generation-related cues and plays an important role in learning robust artifact representations.

\subsubsection{Intra-group Contrasting in SSGC Phase} 
\noindent We further study the contribution of intra-group contrastive learning by removing $\mathcal{L}_{C}$ and optimizing only the disentanglement loss $\mathcal{L}_{D}$ and classification loss $\mathcal{L}_{CLS}$ on $\boldsymbol{f}_{AV}$. As shown in Table~\ref{tab:ablation}, removing $\mathcal{L}_{C}$ consistently weakens the full EDSC framework. For XCLIP-B-FT, recall/F1/AP decrease from 0.8513/0.8122/0.8520 to 0.8427/0.7856/0.8267. For DeMamba-XCLIP-B-FT, the corresponding scores drop from 0.8712/0.8651/0.8760 to 0.8615/0.8342/0.8548. These results show that classification supervision alone is insufficient to fully exploit the semantic grouping structure of SAVGE. By contrasting real and fake videos within the same semantic group, $\mathcal{L}_{C}$ explicitly enlarges the inter-category margin under matched content conditions, helping $\boldsymbol{f}_{AV}$ capture more discriminative generation artifacts. Nevertheless, the degradation caused by removing $\mathcal{L}_{C}$ is generally smaller than that caused by removing $\mathcal{L}_{D}$, suggesting that intra-group contrasting mainly complements feature disentanglement by further refining category separation in the artifact feature space.

\subsubsection{Two-phase Training} 
\noindent Finally, we investigate whether the proposed alternating optimization strategy is necessary by jointly optimizing $\mathcal{L}_{SEA}$ and $\mathcal{L}_{SSGC}$ in a single stage. As shown in Table~\ref{tab:ablation}, joint training leads to severe performance degradation for all three backbones. For XCLIP-B-FT, recall/F1/AP drop to 0.5395/0.4678/0.3916, which is not only far below the full EDSC result of 0.8513/0.8122/0.8520 but also substantially worse than the baseline without EDSC. The same trend appears for DeMamba-XCLIP-B-FT, where joint training obtains only 0.5985/0.5037/0.4668, compared with 0.8712/0.8651/0.8760 for the full EDSC framework. These results indicate that semantic extraction and semantic suppression should not be optimized simultaneously from the beginning. Without a dedicated SEA phase to first establish a reliable semantic anchor, the semantic and artifact branches may interfere with each other, making the suppression objective unstable and preventing $\boldsymbol{f}_{AV}$ from learning meaningful generation-related cues. Therefore, the alternating two-phase strategy \rev{helps stabilize} disentanglement and \rev{improve the performance} of EDSC.

Overall, the ablation results obtained with BERT+TALL~\cite{TALL}, XCLIP-B-FT~\cite{XCLIP}, and DeMamba-XCLIP-B-FT~\cite{DeMamba} show a consistent trend: all core components of EDSC contribute positively across different backbone architectures. Although the magnitude of improvement varies across models, feature disentanglement, intra-group contrasting, and the two-phase training strategy each provide consistent performance benefits. These observations further indicate that the effectiveness of EDSC is attributed to the generality of the training mechanism rather than to a property of any particular backbone.

\begin{figure*}[]
    \centering
    \includegraphics[width=\textwidth]{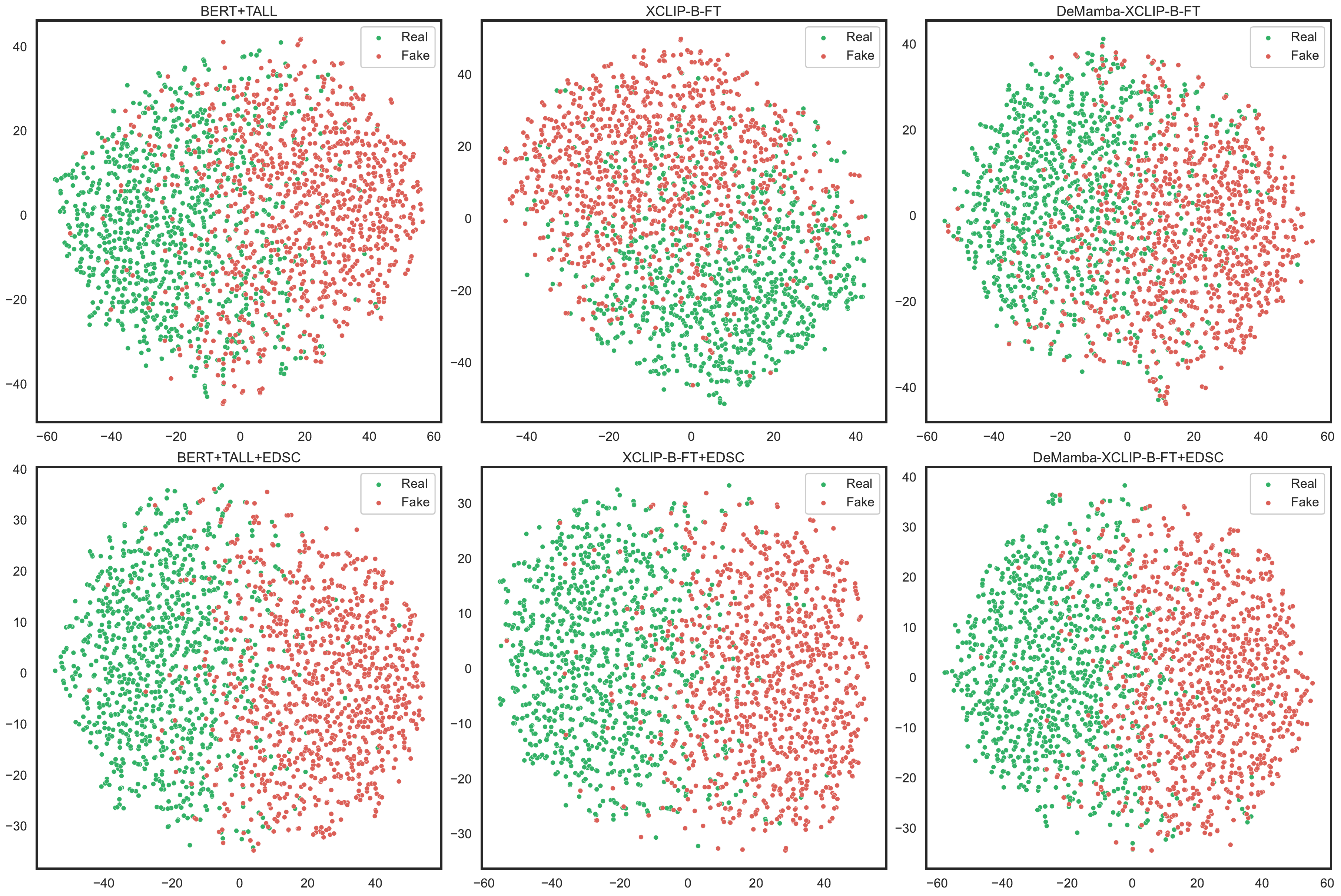}
    \caption{\label{fig:tsne} t-SNE visualization of the learned feature representations produced by BERT+TALL, XCLIP-B-FT, and DeMamba-XCLIP-B-FT, together with their EDSC-enhanced counterparts, on the SAVGE test set.}
\end{figure*}

\subsection{Qualitative Results}

Figure~\ref{fig:tsne} presents t-SNE visualizations of feature representations learned by different baseline detectors and their EDSC-enhanced counterparts on the SAVGE test set. A clear trend can be observed across all three backbones: after incorporating EDSC, the feature distributions of real and fake videos become more compact within each category and more separable across categories. This suggests that EDSC guides the encoders toward representations that better capture generation-related discrepancies rather than being dominated by semantic content variation.

More specifically, the original feature spaces produced by BERT+TALL~\cite{TALL}, XCLIP-B-FT~\cite{XCLIP}, and DeMamba-XCLIP-B-FT~\cite{DeMamba} exhibit noticeable overlap between real and fake samples, indicating that the learned representations still contain ambiguous components that hinder reliable discrimination. In contrast, the EDSC-enhanced versions form more distinct clusters with clearer category boundaries. This observation is consistent with the design of EDSC, which explicitly suppresses semantic information in the artifact branch and strengthens intra-group discrimination through semantic-aware contrasting.

The visualization also shows that the benefit of EDSC is not limited to a specific backbone architecture. Although the degree of separation varies across models, the same qualitative improvement is consistently observed after applying EDSC, demonstrating its compatibility with different detector designs. These results provide intuitive evidence that the proposed framework learns more discriminative representations, helping explain the quantitative gains reported in Table~\ref{tab:main}.

\section{Discussion on Broader Impacts}

\noindent AI-generated videos offer substantial value for creative production, education, and entertainment, but they also introduce risks related to misinformation, impersonation, and the erosion of trust in digital media. Robust detection methods are therefore important for supporting content authenticity and the responsible deployment of generative video technologies.
In this work, SAVGE and EDSC are designed to advance AIGV detection by enabling the study of artifact-aware and semantically grounded representations. In particular, the semantic alignment provided by SAVGE may facilitate the development of detectors that generalize better across generators and application scenarios.

Related concerns are especially important in medical AI, where recent work has surveyed agentic clinical systems, established unified benchmarks for multimodal medical multi-agent systems, and developed evidence-aware distillation for medical vision--language models~\cite{hu2025landscape,qian2026medmaslab,qian2026med}. Although these applications differ from media forensics, they similarly underscore the need to ground model decisions in relevant multimodal evidence and to evaluate robustness across domains.

However, we acknowledge the dual-use nature of this research. Although the dataset and methodology are intended to strengthen detection, they may also inform efforts to evade existing detectors. For this reason, we advocate responsible data release and transparent usage policies.

In the long term, AIGV detection should be viewed as one component of a broader governance framework that also includes platform policy, public awareness, and human oversight. Care is especially needed in deployment, since false positives may affect legitimate content and creators. We hope this work supports future research on trustworthy and responsible synthetic media ecosystems.

\section{Conclusion}

\noindent In this paper, we introduce SAVGE, a large-scale dataset for AI-generated video (AIGV) detection. By constructing semantically aligned real--fake video pairs and providing textual descriptions as semantic anchors, SAVGE offers rich supervisory signals for learning robust and semantically grounded AIGV detectors. Building on this dataset, we further propose Explicit Disentangling and Semantic-aware Contrasting (EDSC), a model-agnostic framework that learns artifact-focused and more generalizable representations. Specifically, EDSC leverages the textual annotations in SAVGE to explicitly disentangle semantically invariant artifact features while using the provided grouping information to enforce a clearer separation between real and fake categories. Extensive experiments on both SAVGE and GenVideo demonstrate the effectiveness of EDSC and show overall improvements in robustness and cross-dataset generalization. We hope that SAVGE and EDSC will serve as useful resources for advancing future research in AIGV detection.

\backmatter

\section*{Funding}
\noindent This work was supported by the National Natural Science Foundation of China (NSFC) under Grant 62476260 and the Fundamental Research Funds for the Central Universities under Grant WK2100000057.

\section*{Author Contributions}
\noindent All authors contributed to this work and the preparation of the manuscript.

\section*{Conflict of Interest}
\noindent All the authors declare that they have no conflict of interest.

\section*{Data Available}
\noindent SAVGE uses real videos and textual annotations from ActivityNet Captions (\url{https://cs.stanford.edu/people/ranjaykrishna/densevid/}), Youku-mPLUG (\url{https://github.com/X-PLUG/Youku-mPLUG}), and Something-Something V2 (\url{https://www.qualcomm.com/developer/software/something-something-v-2-dataset}). GenVideo, used for cross-dataset evaluation, is provided through the authors' repository (\url{https://github.com/chenhaoxing/DeMamba}). Inquiries about access to SAVGE should be directed to the corresponding author, Jiawei Liu (\href{mailto:jwliu6@ustc.edu.cn}{jwliu6@ustc.edu.cn}).

\setlength{\bibsep}{0.8em}
\bibliography{ref}

\end{multicols}
\end{document}